\documentclass[10pt]{article} 

\usepackage[accepted]{tmlr}

\usepackage{hyperref}
\usepackage{url}

\usepackage{bm}
\usepackage{amsmath}
\usepackage{amssymb}
\usepackage{natbib}
\usepackage{graphicx}
\usepackage{subcaption}
\usepackage{multicol}

\usepackage{subcaption}
\usepackage{cleveref}

\usepackage{booktabs}
\usepackage{nicefrac}

\usepackage{algorithm}
\usepackage[beginComment=//~, endComment=~]{algpseudocodex}

\algblock{ParFor}{EndParFor}
\algnewcommand\algorithmicparfor{\textbf{parfor}}
\algnewcommand\algorithmicpardo{\textbf{do}}
\algnewcommand\algorithmicendparfor{\textbf{end\ parfor}}
\algrenewtext{ParFor}[1]{\algorithmicparfor\ #1\ \algorithmicpardo}
\algrenewtext{EndParFor}{\algorithmicendparfor}

\title{Computing distances and means on manifolds with a metric-constrained Eikonal approach}

\author{\name Daniel Kelshaw
		\email djk21@ic.ac.uk\\
		\addr Department of Aeronautics, Imperial College London, London, UK 
		\AND
		\name Luca Magri
		\email l.magri@imperial.ac.uk\\
		\addr Department of Aeronatics, Imperial College London, London, UK\\ The Alan Turing Institute, British Library, London, UK \\ Politecnico di Torino, DIMEAS, Torino, Italy}

\newcommand\restr[2]{{
  \left.\kern-\nulldelimiterspace 
  #1
  \vphantom{|}
  \right|_{#2}
}}

\DeclareMathOperator*{\argmin}{arg\,min}
\DeclareMathOperator*{\grad}{grad}

\begin{document}

\maketitle

\begin{abstract}
Computing distances on Riemannian manifolds is a challenging problem with numerous applications, from physics, through statistics, to machine learning. In this paper, we introduce the {\it metric-constrained Eikonal solver} to obtain continuous, differentiable representations of distance functions (geodesics) on manifolds. The differentiable nature of these representations allows for the direct computation of globally length-minimising paths on the manifold. We showcase the use of metric-constrained Eikonal solvers for a range of manifolds and demonstrate the applications. First, we demonstrate that metric-constrained Eikonal solvers can be used to obtain the Fr\'echet mean on a manifold, employing the definition of a Gaussian mixture model, which has an analytical solution to verify the numerical results. Second, we demonstrate how the obtained distance function can be used to conduct unsupervised clustering on the manifold -- a task for which existing approaches are computationally prohibitive. This work opens opportunities for distance computations on manifolds. 
\end{abstract}


\section{Introduction}
\label{sec:introduction}

Many high-dimensional systems are can be expressed on embedded, low-dimensional manifolds~\citep{fefferman2016TestingManifoldHypothesisa}. Standard Euclidean measures of distance are insufficient for providing meaningful measures of similarity, and thus distance measures which take into account the geometry of the underlying manifold are necessary. Distance computations on nonlinear manifolds are crucial in a range of areas, ranging from statistics to machine learning.
Computing statistics, or conducting reduced-order modelling on manifolds are notable examples of the importance of distance functions on manifolds~\citep{guigui2023IntroductionRiemannianGeometry, pennec2006IntrinsicStatisticsRiemannian}. Tangent principal component analysis (TPCA) aims to find a mean on the manifold, project points from the manifold to the tangent space at the mean, before finally computing PCA on the tangent space~\citep{huckemann2006PrincipalComponentAnalysis, zhang2004principal}. In computing the mean of points on the manifold, the arithmetic mean does not account for the geometry of the manifold, and as a result may not lie on the surface of the manifold. The mean instead must be defined as the point which minimises the square distance to samples on the manifold, known as the Fr\'echet mean. This notion of distance must adhere to the geometry of the manifold.
Applications of distance functions on manifolds extend to the field of machine learning, where embedding methods are prevalent~\citep{wang2014generalized}. Contrastive language-image pre-training (CLIP) make use of a cosine similarity to predict image, text pairings in a zero-shot learning~\citep{radford2021LearningTransferableVisual}. The advent of retrieval-augmented generation in large language modelling makes use of vector databases of text embeddings to search for similar representations, which primarily use measures of Euclidean distance, or cosine similarities~\citep{lewis2021RetrievalAugmentedGenerationKnowledgeIntensive, gao2023retrieval}.
The field of physics holds numerous examples of where manifold-based distances are important. The most notable is that of general relativity. General relativity describes how matter can influence spacetime, as evidenced by gravitational lensing. Due to the curvature of spacetime, the distance that light travels cannot be described through standard Euclidean norms, but rather the notion of geodesic distance is required~\citep{wald2010general}. \\

Much of the existing work on computing geodesic distances on nonlinear manifolds is framed in a discrete manner. Methods typically fall into one of two approaches: {\it (i)} Eikonal-based approaches, and {\it (ii)} length-minimising approaches. 
Distance functions are characterised by having a gradient of unit magnitude everywhere in the domain. The Eikonal equation~\citep{evans2010PartialDifferentialEquations} naturally models this behaviour and, as a result, many methods for computing distance leverage an Eikonal-based approach. The Eikonal equation can be treated as a stationary boundary value problem, or transformed to a time-dependant problem which models arrival time, or wave front propagation. A common means of computing geodesic distances is through the fast marching method, which is a level-set method~\citep{kimmel1998ComputingGeodesicPaths, sethian1999FastMarchingMethods}. The fast marching method treats the problems as a stationary boundary value problem, using a numerical approach to solve a system of equations after discretisation.  This fast marching method is similar to Djikstra's algorithm \citep{dijkstra1959NoteTwoProblems}, which propagates information from a known solution, the boundary values, in a sequential manner -- updating a single grid point at a time so that the solution is strictly increasing. The algorithm used to achieve this comprises two steps: 
{\it (i)} a local solver, used to approximate the gradient at a point and estimate distance to surrounding nodes; and 
{\it (ii)} an update step, responsible for selecting the closest point in order to propagate the solution. The quality of the solution depends on the accuracy of the local solver. 
In a similar vein, there exist fast-sweeping methods for obtaining numerical solutions for the Eikonal equation on a rectangular grid. Fast-sweeping methods use upwind differencing for discretisation, in conjunction with Gauss-Seidel iterations with alternating sweeping ordering~\citep{zhao2005fast, zhao2007parallel}. Ultimately, this allows for parallel updates which reduce the computational complexity of the approach.
Another class for computing geodesic distances are heat methods, which leverage a time-dependant solution approach~\citep{crane2013GeodesicsHeatNew, lin2014GeodesicDistanceFunction, varadhan1967BehaviorFundamentalSolution}. These methods work by {\it (i)} integrating the scalar heat flow for a fixed time; {\it (ii)} computing and normalising the gradient of the resulting scalar field; and 
{\it (iii)} solving a Poisson equation to obtain the closest scalar potential, which is the distance. Again, this approach requires discretisation, and relies on a number of assumptions and approximations. The initial condition for the heat flow requires discrete modelling of a point source, and the heat flow assumption states that the computed distance approaches the true distance as the integration time tends to zero. Finally, machine learning approaches to solving the Eikonal equation have been presented by~\citep{gropp2020ImplicitGeometricRegularization}, but operate under Euclidean assumptions. These assumptions make them unsuitable for operating on nonlinear manifolds. Though machine learning approaches seek an approximate solution, certain guarantees can be made through the choice of the architecture. The authors of \citep{grubas2023NeuralEikonalSolver} model caustics with solutions to the Eikonal equation and bound solutions based on analytical models. \\

In contrast with Eikonal-based approaches, the distance between two points can be computed via a length-minimising curve approach. These shortest curves are known as {\it geodesics} and satisfy the geodesic equation. The geodesic equation, however, describes a class of locally length-minimising curve -- with the desired curve being the shortest curve in this class. Rather than solving a boundary value problem to satisfy the equation, methods instead opt to parameterise a curve between two points, optimising parameters of this curve to yield the globally length-minimising curve.
Curves used to represent geodesics must be at least two times differentiable to satisfy the geodesic equation; thus, cubic splines are a natural choice. Splines are functions which are defined by piecewise polynomials, constructed in such a manner to guarantee continuity between each constituent spline. In addition, boundary conditions can be enforced through construction. In general, splines are parameterised by the locations of the intersections of the piecewise polynomials, referred to as knots. In obtaining a length-minimising geodesic, the location of these knots is optimised to yield a curve with minimising length~\citep{stochman2021github, ni2023ShapeAnalysisComputing, rajkovic2023GeodesicsSplinesEmbeddings}.
Instead of enforcing the structure of cubic splines on the geodesic, another approach is to parameterise a geodesic between two points through a neural network~\citep{chen2018MetricsDeepGenerative}. This neural representation provides a continuous, differentiable representation which can then be integrated to compute the distance. The parameters of the network are optimised to obtain a length-minimising representation of the curve. \\

In this paper, we provide a method to yield a continuous, differentiable representation of the distance function on a nonlinear manifold. Instead of obtaining a distance field around a fixed point, as in the numerical approaches, or only considering a distinct pair of points, as in the length-minimising curve approaches, our method yields a distance function defined directly on the manifold. We further show how we can leverage the differentiable representation of the distance function to obtain globally length-minimising curves between points without incurring the cost of solving additional sub-optimisation problems.\\

We first provide a background of material required to specify the Eikonal equation on the manifold in~§\ref{sec:review}. In §\ref{sec:methodology}, we introduce the  metric-constrained Eikonal equation, which provides a continuous, differentiable distance function on the manifold. Our proposed approach allows us to obtain globally length-minimising paths directly. In~§\ref{sec:results} we showcase results for three manifolds of interest: Euclidean space, the hypersphere, and a non-convex manifold with multiple local extrema. We demonstrate the efficacy of our approach through comparing results to ground-truth solutions, and the recovery of length-minimising paths. Finally, we showcase the applications of the metric-constrained Eikonal solver to compute the Fr\'echet mean on the surface of a Gaussian mixture model, and to perform  unsupervised clustering on curved manifolds. We end the paper with conclusions  in~§\ref{sec:conclusion}. We provide access to our open source library on Github.\footnote{All code is available on GitHub: \url{https://github.com/magrilab/riemax}}


\section{Review on Differential Geometry} \label{sec:review}

Differential geometry provides tools for the study of smooth manifolds, of which Euclidean spaces are a subset. Many real world data lives in non-Euclidean spaces, and so we adopt a non-Euclidean toolbox to deal with this. To provide a motivating example, consider two points $p, q \in \mathbb{S}^2$ on the unit sphere. Using the Euclidean $L_2$ norm provides a distance between these points, representing the length of a line interior to the surface. Should we wish to compute the length of the shortest line which lies on the surface of the sphere, we need to consider a different form of distance metric. In this section, we provide a brief overview of concepts from differential geometry which form the building blocks of our proposed methodology. For a comprehensive explanation of the subject, we refer the reader to ~\citep{lee2018IntroductionRiemannianManifolds, carmo2013RiemannianGeometry}. Notation employed follows the Einstein summation convention.

\subsubsection{Riemannian manifolds and inner product}
A Riemannian manifold~\citep{lee2018IntroductionRiemannianManifolds, carmo2013RiemannianGeometry} is a pair $(M, g)$, where $M$ is a smooth manifold~\citep{lee2012IntroductionSmoothManifolds}, and $g$ is a Riemannian metric on $M$. This metric constitutes a symmetric bilinear form $g: T_pM \times T_pM \rightarrow \mathbb{R}$, where $T_p M$ denotes a vector in the tangent space at $p\in M$. Components of the metric are defined as $g_{ij}(p) = \langle \restr{\partial_i}{p}, \restr{\partial_j}{p} \rangle$, where $\partial_i = \nicefrac{\partial}{\partial x^i}$. For arbitrary vectors $v, w \in T_p M$, the metric allows for computation of the inner product on the tangent space~\citep{lee2018IntroductionRiemannianManifolds}
\begin{equation} \label{eqn:review:inner_product}
  \langle v, w \rangle_g = g_p(v, w) = g_{ij} v^i w^j.
\end{equation}
The definition of the inner product allows us to compute distances and angles in the tangent space at a point $p \in M$; ultimately, providing the building blocks for operating on Riemannian manifolds. In Euclidean geometry, the metric is defined as ${\tilde g}_{ij} = \delta_{ij}$, which is the zero-curvature condition. Using this metric in conjunction with the definition of the inner product, we recover the standard definition of the inner product on Euclidean spaces, $\langle v, w \rangle_{\tilde g} = v^i w^i$. Every smooth manifold admits a Riemannian metric~\citep{lee2018IntroductionRiemannianManifolds}.

\subsection{Induced metric}
In order to make the problem computationally possible, we work with embedded submanifolds. Suppose $(\tilde{M}, \tilde{g})$ is a Riemannian manifold, and $M \subseteq \tilde{M}$ is an embedded submanifold. Given a smooth immersion $\iota: M \hookrightarrow \tilde{M}$, where $\dim(M) \leq \dim(\tilde{M})$, the metric $g = \iota^\ast \tilde{g}$ is referred to as the metric induced by $\iota$, where $\iota^\ast$ is the pullback~\citep{lee2018IntroductionRiemannianManifolds}. If $(M, g)$ is a Riemannian submanifold of $(\tilde{M}, \tilde{g})$, then for every $p \in M$ and $v, w \in T_p M$, the induced metric is (from Nash embedding theorems)
\begin{equation} \label{eqn:induced_metric}
  g_p \left( v, w \right) 
  = \left( \iota^\ast \tilde{g} \right)_p \left( v, w \right)
  = \tilde{g}_{\iota(p)} \left( d \iota_p(v), d \iota_p(w) \right).
\end{equation}
This equivalence forms the guiding principle that allows us to compute induced metrics induced by differentiable immersions~\citep{carmo2013RiemannianGeometry}. We leverage this idea throughout the paper to numerically define metrics.

\subsection{Covariant differentiation and parallel transport}
Given the inner product on a manifold, we can introduce the notion of covariant differentiation; a  means of specifying derivatives along tangent vectors. For a vector $w \in T_p M$ and a vector field $v \in TM$, the covariant derivative is defined as~\citep{carmo2013RiemannianGeometry}
\begin{equation} \label{eqn:covariant_derivative}
  \nabla_v w 
    = v^j \nabla_{\partial_j} w^i \partial_i 
    = (v^j w^k_{\phantom{k};j}) \partial_k 
    = \left( v^j w^k_{\phantom{k},j} + \Gamma^k_{\phantom{k}ij} v^j w^i \right) \partial_k,
\end{equation}
where, for brevity, we introduce the notation $w^k_{\phantom{k},j}, w^k_{\phantom{k};j}$ to denote the partial, and covariant derivative respectively -- each taken with respect to the $j$th coordinate vector field. The parenthesised expression in Eq.~\eqref{eqn:covariant_derivative} provides an intuitive explanation for the concept of covariant differentiation. The first term represents the directional derivative, and the second term is responsible for quantifying the twisting of the coordinate system. The term $\Gamma^k_{\phantom{k}ij}$ is referred to as the Christoffel symbols of the second kind, or the affine connection~\citep{lee2018IntroductionRiemannianManifolds}, which provide a measure of how the basis changes over the manifold. The Christoffel symbols are defined as
\begin{equation}
  \Gamma^k_{\phantom{k}ij}
    = \partial_j(\partial_i) \cdot \partial^k 
    = \tfrac{1}{2} g^{km} \left( g_{mi,j} + g_{mj,i} - g_{ij,m} \right),
\end{equation}
where $\partial^k = g^{km} \partial_{m}$ is the index-raised coordinate vector field (contravariant), and $g^{km}$ is the inverse of the metric tensor, such that $\delta^{i}_{\phantom{i}k} = g^{ij} g_{jk}$. In general, transporting a vector $w \in T_p M$ along the manifold does not guarantee that it remains parallel to itself in the tangent planes along the path. Definition of the covariant derivative provides a sufficient condition to ensure parallelism is conserved along a trajectory. Mathematically this is expressed as
\begin{equation} \label{eqn:geodesic_equation}
  \nabla_v w =  v^j w^k_{,j} + \Gamma^k_{ij} v^j w^i = 0.
\end{equation}
This condition for parallelism can be exploited to extend the notion of straight curves to the manifold. Geodesics, which are straight lines in Euclidean space, allow us to define a notion of distance.

\subsection{Geodesics}
A geodesic $\gamma: [a, b]\subset \mathbb{R} \rightarrow M$ generalises the notion of a straight line on the manifold. These curves are locally length-minimising because they constitute a solution of the Euler-Lagrange equations, which, mechanically, describe the motion of a rigid body devoid of acceleration~\citep{carmo2013RiemannianGeometry}. Leveraging the definition of covariant differentiation, we can describe a geodesic as a curve for which tangent vectors remain parallel to themselves as they are transported along the curve, yielding the geodesic equation~\citep{lee2018IntroductionRiemannianManifolds}
\begin{equation}
  \nabla_{\dot{\gamma}} \dot{\gamma} = \ddot{\gamma}^k + \Gamma^k_{\phantom{k}ij} \dot{\gamma}^i \dot{\gamma}^j = 0,
  \qquad \text{s.t.} \quad 
  \gamma(\lambda = 0) = \gamma_0, \; \dot{\gamma} (\lambda = 0) = \dot{\gamma}_0, \; \lambda \in [0, 1]
\end{equation}
where derivatives $\ddot{\gamma}, \dot{\gamma}$ are taken with respect to the affine parameter $\lambda$. This geodesic equation ensures the speed $\langle \dot{\gamma}, \dot{\gamma} \rangle^{0.5}_g$ is constant along the trajectory, allowing us to classify the  parameterisations $\lambda \in [0, 1]$ as unit-speed geodesics. Conversely, the geodesic can be parameterised with respect to length to yield a unit-distance geodesic~\citep{lee2018IntroductionRiemannianManifolds}. \\

The definition of the geodesic equation gives rise to two mappings of interest. Given a point $p \in M$, the $\exp_p$ and $\log_p$ maps are, respectively
\begin{align}
  \exp_p &: T_p M \rightarrow M, \\
  \log_p &: M \rightarrow T_p M.
\end{align}
The $\exp_p$ map defines an initial value problem. Given a vector $v \in T_p M$, there exists a unique geodesic $\gamma: [0, 1] \rightarrow M$ satisfying $\gamma(0) = p, {\dot \gamma}(0) = v$. The exponential map is then defined as $\exp_p(v) = \gamma(1)$. We can define the $\log_p$ map as the functional inverse of the $\exp_p$ map, that is, $\exp_p \circ \log_p : q \mapsto q \; \forall q \in M$. The $\log$ map, however, is not necessarily unique because there may be many $v \in T_p M$ for which $\exp_p v = q$, providing a challenge in defining distance on the manifold. In §\ref{sec:methodology}, we overcome this by defining the $\log^\ast_p$ map, which makes use of the gradient of the distance field to obtain a unique solution.

\subsection{Distances on manifolds}
The notion of distance on the manifold is defined by geodesics. Although geodesics are length-minimising, this is only in a local sense. Given two points $p, q \in M$, the geodesic distance is defined as the infimum of the length of all valid geodesics~\citep{carmo2013RiemannianGeometry} 
\begin{equation} \label{eqn:review:distance}
  d_g\left( p, q \right) = \inf_\gamma \left\{ \int_0^1 \langle \dot{\gamma}(t), \dot{\gamma}(t) \rangle^{\nicefrac{1}{2}}_{\gamma(t)} dt : \gamma(0) = p, \gamma(1) = q \right\}.
\end{equation}
A geodesic $\gamma$ is said to be minimising \textit{iff} there exist no shorter, valid geodesics with the same endpoints. This forms the basis for approaches which seek to obtain length-minimising curves between two points, as outlined in~§\ref{sec:introduction}.

\subsection{Curvature}
The Riemann curvature tensor $R^{l}_{\phantom{l}ijk}$ provides a means to quantify curvature on the manifold; a Riemannian manifold with zero curvature is referred to as {\it flat} and is locally isomorphic to Euclidean space. Given two {\it infinitesimally close} geodesics, curvature is responsible for geodesic deviation. To obtain a scalar value for the curvature, one can successively contract over indices of the Riemann curvature tensor~\citep{lee2018IntroductionRiemannianManifolds}
\begin{equation} \label{eqn:curvature_progression}
  R^l_{\phantom{l}ijk} = 
    \Gamma^l_{\phantom{l}ik,j}- \Gamma^l_{\phantom{l}ij,k} + \Gamma^l_{\phantom{l}jm} \Gamma^m_{\phantom{m}ik} - \Gamma^l_{\phantom{l}km} \Gamma^m_{\phantom{m}ij}
  \quad \rightarrow \quad
  R_{ij} = R^{m}_{\phantom{m}imj}
  \quad \rightarrow \quad
  R = g^{ij} R_{ij},
\end{equation}
where $R^l_{\phantom{l}ijk}$ denotes the Riemann curvature tensor; $R_{ij}$ the Ricci curvature tensor, obtained by contracting  over the first and third indices; and $R$, the Ricci scalar -- the geometric trace of the Ricci tensor with respect to the metric. The Ricci scalar is a local invariant on the manifold and assigns a single real number to the curvature at a particular point. The Ricci scalar relies solely on the definition of the metric tensor $g_{ij}$. \\ 

In this paper, we provide a computational framework for operating on differentiable manifolds. We leverage the automatic differentiation capabilities of \texttt{jax}~\citep{jax2018github} to compute the metric induced by smooth immersions as described in Eq.~\eqref{eqn:induced_metric}, as well as all derivative properties.


\section{The metric-constrained Eikonal solver and methodology} \label{sec:methodology}

Given a smooth immersion $\iota: M \hookrightarrow \mathbb{R}^n$, with $\dim(M) \leq n$, we propose a method to obtain a continuous, differentiable representation of the distance function on $M$. In this section, we first discuss properties of distance functions and metric spaces, providing criteria that our proposed distance function should satisfy. Second, we build intuition for a distance function constrained around a {\it single-point} fixed in the domain. Specifically, we {\it (i)} introduce the Eikonal equation as a means to express distance functions; {\it (ii)} outline a principled means by which to tractably compute $\log_p$ maps; and {\it (iii)} deploy neural networks to parameterise the distance function. Third, we extend the approach to compute generic and global distance functions. Finally, we provide a brief outline of network architectures and training techniques. \\ 

\subsection{Properties of distance functions}
A distance function $d: M \times M \rightarrow \mathbb{R}$ is a function with a gradient of unit magnitude at every point in the domain. Defining a distance function on the manifold is equivalent to defining a metric space $(M, d)$~\citep{lee2012IntroductionSmoothManifolds}, which for all points $p, q, r \in M$ satisfy
\begin{equation}
\begin{aligned}[c]
  &\textbf{M1:} \quad d(q, q) = 0 \\
  &\textbf{M2:} \quad d(p, q) \geq 0
\end{aligned}
\qquad \qquad
\begin{aligned}
  &\textbf{M3:} \quad d(p, q) = d(q, p) \\
  &\textbf{M4:} \quad d(p, q) \leq d(p, r) + d(r, q)
\end{aligned}
\end{equation}
In other words, {\bf M1} defines the notion of {\it zero distance}; {\bf M2} represents {\it positivity}; {\bf M3} is the notion of {\it symmetry}; and {\bf M4} states the {\it triangle inequality} holds. We can leverage the definition of geodesics to impose an additional property
\begin{equation}
  \textbf{M5:} \quad (\iota^\ast d_E)(p, q) \leq d(p, q),
\end{equation}
which states that the geodesic distance is not shorter than the Euclidean distance in the image space of the immersion. This follows immediately from acknowledging that geodesics are length-minimising curves on the manifold. Our proposed methodology ensures that {\bf M}$\{{\bm 1},{\bm 2},{\bm 3},{\bm 5}\}$ hold by construction.

\subsection{Single-point solutions} \label{sec:methodology:sp_solutions}
We first discuss single-point variations of distance functions. We refer to {\it single-point} solutions as distance functions with a fixed point $p \in M$, such that $d_p : q \mapsto d(p, q)$. This allows us to introduce the methodology, before extending the approach to handle global distance functions in §\ref{sec:methodology:global_solutions}. \\ 

\subsubsection{Eikonal equations}
We consider a class of nonlinear first-order partial differential equations, the Eikonal equations, for which the solution satisfies the properties of the distance function, i.e., the magnitude of the gradient is unity at all points in the domain. Fixing a point $p \in M$, the distance $\varphi: q \mapsto d(p, q)$ satisfies the equation
\begin{equation}
  \lVert \nabla \varphi \rVert = 1
  \quad
  \text{s.t.}
  \quad
  \restr{\varphi}{p} = 0.
\end{equation}
On a Riemannian manifold $(M, g)$, we  augment the formulation with the exterior derivative, which we refer to as {\it the metric-constrained Eikonal equation}
\begin{equation}
  \lVert \nabla \varphi \rVert
  = \langle \grad \varphi, \grad \varphi \rangle_g
  = \varphi^{,i} \varphi_{,i} = 1
  \quad
  \text{s.t.}
  \quad
  \restr{\varphi}{p} = 0,
\end{equation}
where $(\grad \varphi)^i = g^{ij} \varphi_{,j}$ is the exterior derivative, and $\varphi^{,i} = g^{ij} \varphi_{,j}$. Taking derivatives in this way ensures that the solution adheres to the geometry of the manifold, as described by the metric induced by the immersion.

\subsubsection{Implicit $\log$ maps} \label{sec:methodology:implicit_log_map}
Consider two points $p, q \in M$, and level-sets of constant distance around the point $p$. The shortest path connecting $p, q$ must travel orthogonal to the level-sets, that is, geodesics are defined by the gradient of the distance function. Precisely, the exterior derivative of the distance function produces the geodesic flow, which is a section of the tangent bundle. This flow satisfies the geodesic equation~\citep{carmo2013RiemannianGeometry},
\begin{equation}
  \nabla_{\grad \varphi} \grad \varphi = 0,
\end{equation}
This geodesic flow allows us to obtain globally length minimising geodesics. Although the $\log_p$ map is not unique, we formulate a $\log^\ast_p$ map to obtain the tangent vector which yields the globally shortest geodesic between two points. We exploit the obtained distance function and the corresponding geodesic flow to obtain a vector field describing the dynamics of globally length-minimising paths. 
We first consider the system
\begin{equation}
  {\dot \gamma}(\lambda) = - \grad \varphi (\gamma(\lambda)), 
  \qquad \text{with} \quad \gamma: \left[0, \varphi(q; p) \right) \rightarrow M.
\end{equation}
By taking $\gamma(\lambda = 0) = q, \dot{\gamma}(\lambda = 0) = -\grad \varphi(q; p)$ as the initial condition, we can define the $\log^\ast_p$ map as
\begin{equation}
  \log^\ast_p: q \mapsto \lim_{\lambda \to \varphi(q; p)^{-}} \left[ 
    \grad \varphi \left( \gamma(\lambda = 0) + \int_0^{\lambda} {\dot \gamma}(\lambda) d\lambda \right)
  \right].
\end{equation}
Intuitively, given points $p, q \in M$ we compute the $\log^\ast_p$ map by {\it (i)} evaluating the distance, $\varphi(q; p)$; {\it (ii)} starting at $q$, trace the gradient field back to $p$ by integrating the initial value problem along the length of the geodesic; and {\it (iii)} evaluate $\grad \varphi(p)$. We cannot directly evaluate $\grad \varphi(p)$ as the solution is not unique because $p$ is the source of the geodesic flow. Instead, we follow the gradient field in the reverse direction from a point $q$, tracing out the globally length-minimising geodesic which terminates at $p$.
Standard methods for computing the $\log^\ast_p$ map rely on parameterising and optimising curves between pairs of points. By employing a metric-constrained Eikonal solver approach, once the distance function has been obtained, the length-minimising geodesics, and thus the $\log^\ast_p$ map can be computed straightforwardly (§\ref{sec:results}).

\subsubsection{Network parameterisation}
We seek a continuous, differentiable solution to the metric-constrained Eikonal equation. To this end, we parameterise the distance function by a neural network ${\tilde \varphi}_\theta : M \rightarrow \mathbb{R}$. Neural networks are universal function approximators of continuous functions~\citep{hornik1989MultilayerFeedforwardNetworks}. The output of the network is augmented to yield the distance function
\begin{equation}
  \varphi_\theta(q; p)
  = \left( \iota^\ast d_E \right)(p, q) \left[ 1 + (\sigma \circ \tilde{\varphi}_\theta) (q; p) \right],
\end{equation}
where $\sigma \in \mathcal{C}^{\geq2} \left( \mathbb{R}, \mathbb{R}_{\geq 0} \right)$ is strictly increasing and at least twice differentiable; $\circ$ denotes function composition; and $d_E: (p, q) \mapsto \lVert p - q \rVert_2$ is the Euclidean distance function. Defining the distance function in this manner serves three distinct purposes: {\it (i)}, the constraint $\restr{\varphi}{p} = 0$ is handled implicitly ({\bf M1}); the output of the network is guaranteed to be positive ({\bf M2}); and {\it (iii)}, the inequality $\varphi(q; p) \geq (\iota^\ast d_E)(p, q)$ is guaranteed ({\bf M5}). We seek a solution ${\varphi_{\theta^\ast} = d_p(q)}$ through solving the optimisation problem
 for the trainable parameters %
\begin{equation} \label{eqn:optimisation_problem}
  \theta^\ast = \argmin_\theta{\mathcal{L}(\theta; p)}
    \quad \text{where} \;
  \mathcal{L} = \mathbb{E}_{q} \left( \lVert
    {\varphi_\theta}^{,i}(q) {\varphi_\theta}_{,i}(q) - 1
  \rVert^2 \right),
\end{equation}
where the expectation is taken with respect to a distribution $q \sim \mathcal{D}$ in $M$. This optimisation problem seeks to minimise the residual of the metric-constrained Eikonal equation across the domain.

\subsection{Global solutions} \label{sec:methodology:global_solutions}
In this subsection, instead of constraining the solution to the Eikonal equation to a single point, we consider learning a global solution $\varphi: (p, q) \mapsto d(p, q)$. We first introduce an additional equation to enforce the Eikonal constraint on both variables
\begin{equation}
  \lVert \nabla_p \varphi(p, q) \rVert = \lVert \nabla_q \varphi(p, q) \rVert = 1
  \quad \text{s.t.} \quad \restr{\varphi}{p=q} = 0,
\end{equation}
where $\nabla_p, \nabla_q$ denotes partial derivatives with respect to the respective function arguments. As before, we parameterise the solution with a neural network, this time taking points $p, q \in M$ as inputs. As well as the constraints imposed for the single-point case, we also wish to impose symmetricity ({\bf M3}). To achieve this, we augment the output of the network to define the global distance function
\begin{equation} \label{eqn:augmentation:global}
  \varphi_\theta(p, q)
  = \left( \iota^\ast d_E \right)(p, q) \left[ 1 + \sigma\circ \left( \tfrac{1}{2} ( \tilde{\varphi}_\theta (p, q) + \tilde{\varphi}_\theta (q, p) ) \right) \right].
\end{equation}
As a result of this imposed reciprocity, we know the equality $\lVert \nabla_p \varphi(p, q) \rVert = \lVert \nabla_q \varphi(p, q) \rVert$ holds. The symmetricity constraint ensures that we still need only a single equation to specify the global solution to the metric-constrained Eikonal equation~\citep{grubas2023NeuralEikonalSolver}, and so we follow a similar optimisation approach as in the {\it single-point} case described in~§\ref{sec:methodology:sp_solutions}. Thus, we seek a solution ${\varphi_{\theta^\ast} = d(p, q)}$ through solving
\begin{equation} \label{eqn:global:optimisation_problem}
  \theta^\ast = \argmin_\theta{\mathcal{L}(\theta)}
    \quad \text{where} \;
  \mathcal{L} = \mathbb{E}_{q} \left( \lVert
    {\varphi_\theta}^{,i}(p, q) {\varphi_\theta}_{,i}(p, q) - 1
  \rVert^2 \right),
\end{equation}
where the expectation is taken with respect to a distribution $q \sim \mathcal{D}$ in $M$. (The network outputs are augmented in such a manner to provide a lower-bound on the distance based on the pullback of the Euclidean distance, $\iota^\ast d_E$ ({\bf M5}). (We provide an  upper-bound of ther the distance in Appendix~\ref{app:upper_bounds}. This method is not showcased in the results of this paper as it increased computational complexity with no significant impact on the quality of the results.)

\subsection{Network architecture and training} \label{sec:methodology:network}
We parameterise the distance functions $\varphi_\theta$ with a neural network; in particular, we employ the modified multi-layer perceptron (MLP) architecture as described by~\citep{wang2021UnderstandingMitigatingGradient}. This choice of architecture has two main advantages: {\it (i)} the introduction of positional embeddings; and {\it (ii)} the use of residual connections aid in mitigating numerical gradient instability. Given the reliance on our proposed methodology on the use of automatic differentiation for computing the loss, the modified MLP is a natural choice. For an extensive overview on modified MLPs, we refer the reader to~\citep{wang2023ExpertGuideTraining}.

The optimisation of the network ${\tilde \varphi}_\theta$ is performed through a gradient-descent approach, which employs the \texttt{adam} optimiser~\citep{kingma2017AdamMethodStochastic} to minimise the expectation of the loss. The initial learning rate for the optimiser is prescribed in each case, and then follows an exponential decay schedule to aid convergence. The expectation of the loss should provide a meaningful estimate of the residual error, and is taken over a distribution $q \sim \mathcal{D} \in M$. A standard approach would be to sample points in a uniform distribution, selecting all points in the domain with equal probability. However, we can exploit properties of the geometry to provide improved estimates.

\subsubsection{Curvature-based sampling}
The solution $\varphi$ to the metric-constrained Eikonal equation is more sensitive in regions of high curvature, which is a consequence of induced geodesic deviation~\citep{carmo2013RiemannianGeometry}. We mitigate the impact of geodesic deviation by  employing a curvature-based sampling strategy, which ensures that we can capture the solution appropriately in regions of high curvature. Employing the Metropolis-Hastings~\citep{metropolis1953equation, hastings1970MonteCarloSampling} sampling scheme, we draw samples in regions of the manifold that exhibit higher degrees of scalar curvature, which is characterised by the magnitude of the Ricci scalar \citep{lee2018IntroductionRiemannianManifolds}. An outline of this curvature-based sampling is provided in algorithm~\ref{applications:algorithm:sampling}.
\begin{algorithm}
\caption{Metropolis-Hastings curvature-based sampling}\label{applications:algorithm:sampling}
\begin{algorithmic}
\Require $q_0 \in M, \delta \in \mathbb{R}$
\Comment{$q_0$ is the initial condition, $\delta$ controls the step-size}
\For{$t \gets 1, \dots n$}
  \State $q' \gets q_{t - 1} + \delta \mathcal{N}(0, 1)$
  \Comment{propose the location of a new sample}
  \State $u \gets \mathcal{U}(0, 1)$
  \State $\alpha \gets \log(\lvert R(q') \rvert) - \log(\lvert R(q_{t - 1}) \rvert)$
  \Comment{compute acceptance ratio}
  \If{$\log(u) \leq \alpha$}
  \Comment{accept the proposed point}
    \State $q_t \gets q'$
  \Else
  \Comment{reject the proposed point}
    \State $q_t \gets q_{t - 1}$
  \EndIf
\EndFor
\end{algorithmic}
\end{algorithm}

\section{Results} \label{sec:results}

\begin{figure}[htb]
  \centering
  \captionsetup[subfigure]{justification=centering}
  \begin{subfigure}[t]{0.32\textwidth}
    \centering
    \includegraphics[width=\textwidth]{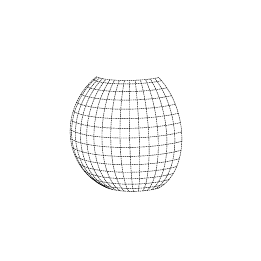}
    \caption{Patch of $\mathbb{S}^2$}
    \label{fig:manifolds:sphere}
  \end{subfigure} \hfill
  \begin{subfigure}[t]{0.32\textwidth}
    \centering
    \includegraphics[width=\textwidth]{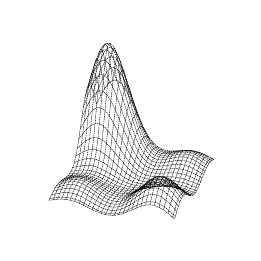}
    \caption{Gaussian mixture model}
    \label{fig:manifolds:gmm}
  \end{subfigure} \hfill
  \begin{subfigure}[t]{0.32\textwidth}
    \centering
    \includegraphics[width=\textwidth]{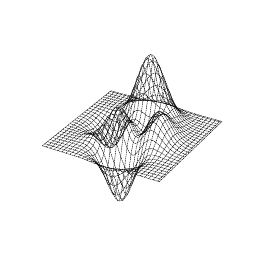}
    \caption{Peaks manifold}
    \label{fig:manifolds:peaks}
  \end{subfigure}
  \caption{Overview of the manifolds used throughout the paper, visualised in $\mathbb{R}^3$. Panel (\subref*{fig:manifolds:sphere}) depicts a patch of the two-sphere, panel (\subref*{fig:manifolds:gmm}) shows a model defined by the probability density function of a Gaussian mixture model, and panel (\subref*{fig:manifolds:peaks}) shows a multi-modal manifold.}
  \label{fig:manifolds}
\end{figure}

In order to demonstrate the efficacy of our proposed methodology, we show results for three manifolds of interest: {\it (i)} Euclidean space, {\it (ii)} the unit hypersphere, and {\it (iii)} the peaks manifold. We first consider the Euclidean case, providing a baseline for which we can verify the method in the absence of curvature. Second, we introduce constant curvature by obtaining distance functions on the unit hypersphere. In each of these cases, there exists an analytical form of the distance function; parameterising the manifolds by the number of dimensions allows us to investigate how the proposed methodology scales with increasing dimensionality. Finally, we investigate the peaks manifold, which is a non-convex manifold with multiple local extrema. There exists no analytical form for the distance function on the peaks manifold, and as a result we show comparisons against the length-minimising approach. An overview of the manifolds used throughout the paper, with the exception of Euclidean space, is provided in Figure~\ref{fig:manifolds} where visualisations are shown in $\mathbb{R}^3$.
This section focuses on demonstrating the proposed methodology, and §\ref{sec:applications} highlights the benefits that using a continuous-differentiable distance function can bring.

\subsection{Euclidean Space} \label{sec:results:euclidean}
We first consider the limiting case of the Euclidean manifold, for which there is zero curvature across the entire domain. As discussed in §\ref{sec:review}, the metric tensor is the identity matrix. Despite the apparent simplicity of the manifold, this provides a foundation for analysis. 
The Euclidean metric is induced by the smooth immersion $\iota: x \mapsto x$ with $x \in \mathbb{R}^n$ (the identity function). The corresponding distance function is the $\ell^2$ norm, $d_g : (p, q) \mapsto \lVert p - q \rVert_2$. Pairing the manifold with the corresponding distance function yields a metric space.

\subsubsection{Performance}
In order to assess the performance of the proposed methodology, we seek to recover the distance function for dimensions $n \in \{2, 5, 10, 15, 20\}$. In each case, we seek a solution $\varphi_{\theta^\ast} = d_g$. Performance is measured using the relative $\ell^2$ error between the predicted, and ground truth distance. We define the error
\begin{equation} \label{eqn:relative_err}
  \varepsilon = \sqrt{ \frac{
    \mathbb{E} [ ( \varphi_\theta (p, q) - d_g (p, q) )^2 ]
    }{
    \mathbb{E} [ d_g (p, q)^2 ]
  } }
  \qquad \text{where} \quad p, q \sim \mathcal{U}_{[M^+, M^-]},
\end{equation}
where $M^+, M^-$ denote the bounds imposed for training on the manifold. For the purposes of this work, we consider a bounded subset of the manifold, namely $M \in [-3, 3] \subset \mathbb{R}^n$, so we can standardise the inputs to the network.

\subsubsection{Optimisation}
We seek a solution to the global Eikonal equation (global optimisation problem Eq.~\eqref{eqn:global:optimisation_problem}). Following the initialisation of the network, optimisation is performed with the \texttt{adam} optimiser. Further details on the network and optimisation parameters can be found in Appendix~\ref{app:training_parameters}. For each case $n \in \{2, 5, 10, 15, 20\}$, we train a total of five models, each initialising weights of the network with a different random seed. Each model is optimised for a total of $5 \times 10^4$ parameter updates, empirically determined to provide sufficient convergence. The loss is computed using using $2^{14}$ points uniformly sampled for each $p, q$ (we do not employ curvature-based sampling as the curvature is uniformly zero across the domain).

\begin{figure}[htb]
  \centering
  \captionsetup[subfigure]{justification=centering}
  \begin{subfigure}[t]{0.32\textwidth}
    \centering
    \includegraphics[width=\textwidth]{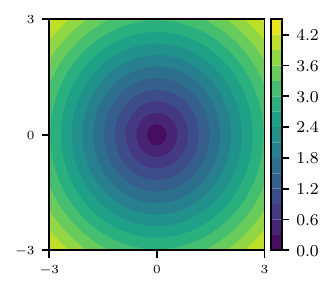}
    \caption{Predicted distance field}
    \label{fig:euclidean:pred_distance}
  \end{subfigure} \hfill
  \begin{subfigure}[t]{0.32\textwidth}
    \centering
    \includegraphics[width=\textwidth]{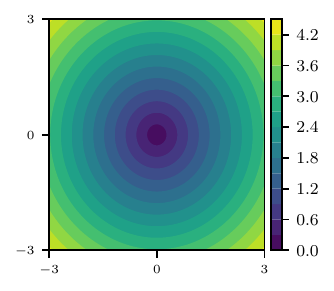}
    \caption{True distance field}
    \label{fig:euclidean:true_distance}
  \end{subfigure} \hfill
  \begin{subfigure}[t]{0.32\textwidth}
    \centering
    \includegraphics[width=\textwidth]{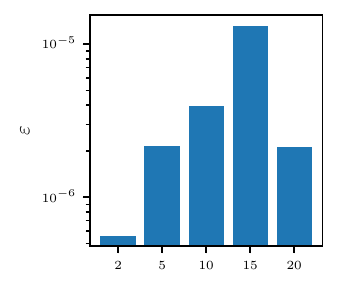}
    \caption{Error study}
    \label{fig:euclidean:dimension_study}
  \end{subfigure}
  \caption{Results for the Euclidean manifold. Panel (\subref*{fig:euclidean:pred_distance}) shows the predicted distance around a point, panel (\subref*{fig:euclidean:true_distance}) shows the corresponding true distance from the point, and panel (\subref*{fig:euclidean:dimension_study}) shows the mean relative $\ell^2$ error across five repeats for increasing dimensions $n \in \{2, 5, 10, 15, 20\}$.}
  \label{fig:euclidean}
\end{figure}

To provide an illustrative example, we first consider the case $\mathbb{R}^2$ for which we can visualise the resulting distance fields. Figures \ref{fig:euclidean}(\subref*{fig:euclidean:pred_distance}, \subref*{fig:euclidean:true_distance}) show the predicted distance, and true distance around the point $p = 0$. The distance field is characterised by level-sets forming concentric circles around the origin. Qualitatively, we observe that the predicted and ground-truth fields are identical, demonstrating that the model is able to recover the distance function appropriately.
Figure \ref{fig:euclidean:dimension_study} provides quantitative results for each of the cases explored, quoting the mean relative $\ell^2$ error achieved across the five runs. The standard deviation for each $n$ did not exceed $6.239 \times 10^{-8}$, and as such error bars are not reported. We observe that, with the exception of $n=20$, the error increases with increasing dimension $n$. This behaviour is expected, as the distance function becomes more challenging to model in higher dimensions.

\subsection{Hypersphere} \label{sec:results:hypersphere}
We continue our investigation by considering a patch of the hypersphere $\mathbb{S}^n$, another manifold for which there exists an analytical distance function. A visualisation is provided in Figure~\ref{fig:manifolds:sphere}. While our previous example of Euclidean space exhibits zero curvature, the hypersphere is characterised by constant curvature, with Ricci scalar $R = n(n - 1)r^{-2}$ where $r$ is the radius~\citep{carmo2013RiemannianGeometry}. Without loss of generality, we consider the unit hypersphere, $r=1$. The introduction of curvature marks a notable challenge for which Euclidean-based methods are not positioned to handle.
An induced metric of the hypersphere manifold is provided by the immersion $\iota: \mathbb{S}^n \rightarrow \mathbb{R}^{n + 1}$. A  definition of the immersion is provided in Appendix~\ref{app:hypersphere}. The distance function on the unit hypersphere is provided by the cosine rule, $d_g : (p, q) \mapsto \arccos \left( \iota(p) \cdot \iota(q) \right)$. This manifold, in conjunction with the associated distance function, provides the metric space.

\subsubsection{Performance}
Similarly to the experiments run in the Euclidean case, we seek to recover the distance function for each $n \in \{2, 5, 10, 15, 20\}$. The performance is measured by the relative $\ell^2$ error between the predicted, and ground-truth distances; employing the definition in Eq.~\eqref{eqn:relative_err}. Imposing a coordinate system on the hypersphere inevitably leads to a singularity which can cause numerical issues. As a result of this singularity, and the periodicity of the manifold, we impose bounds $x \in [\nicefrac{\pi}{6}, \nicefrac{5 \pi}{6}] \subset \mathbb{S}^n$. These bounds are used to standardise inputs to the network.

\subsubsection{Optimisation}
The optimisation scheme used for obtaining a solution to the metric-constrained Eikonal equation in the case of the hypersphere is identical to that used in the Euclidean case, described in §\ref{sec:results:euclidean}. In this case, we also do not employ the curvature-based sampling method, as we know {\it a-priori} that the manifold exhibits constant scalar curvature.

\begin{figure}[htb]
  \centering
  \captionsetup[subfigure]{justification=centering}
  \begin{subfigure}[t]{0.32\textwidth}
    \centering
    \includegraphics[width=\textwidth]{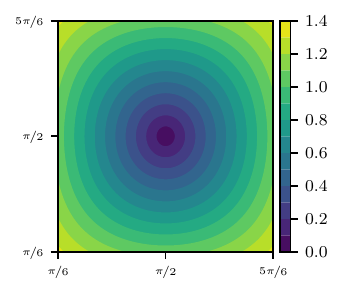}
    \caption{Predicted distance field}
    \label{fig:hypersphere:pred_distance}
  \end{subfigure} \hfill
  \begin{subfigure}[t]{0.32\textwidth}
    \centering
    \includegraphics[width=\textwidth]{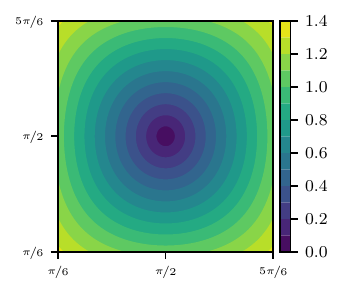}
    \caption{True distance field}
    \label{fig:hypersphere:true_distance}
  \end{subfigure} \hfill
  \begin{subfigure}[t]{0.32\textwidth}
    \centering
    \includegraphics[width=\textwidth]{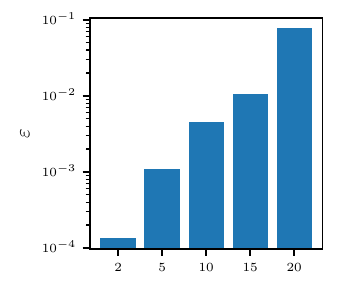}
    \caption{Error study}
    \label{fig:hypersphere:dimension_study}
  \end{subfigure}
  \caption{Results for the hypersphere manifold. Panel (\subref*{fig:hypersphere:pred_distance}) shows the predicted distance around a point, panel (\subref*{fig:hypersphere:true_distance}) shows the corresponding true distance from the point, and panel (\subref*{fig:hypersphere:dimension_study}) shows the mean relative $\ell^2$ error across five repeats for dimensions $n \in \{ 2, 5, 10, 15, 20 \}$.}
  \label{fig:hypersphere}
\end{figure}

For the case of $\mathbb{S}^2$, we provide a visualisation of the distance function around the point $p=\nicefrac{\pi}{2}$. Results in Figures~\ref{fig:hypersphere}(\subref*{fig:hypersphere:pred_distance}, \subref*{fig:hypersphere:true_distance}) show the predicted distance, and true distance fields respectively. The presence of curvature, though subtle, is apparent in these examples, with the level-sets of distance appearing warped when compared with the original Euclidean case. Despite the presence of curvature, the predicted and true distance fields show good agreement.
Figure \ref{fig:hypersphere:dimension_study} shows the results for recovering the distance function on the hypersphere for each of the cases, quoting the mean relative $\ell^2$ error achieved across the five runs. The standard deviation for each case did not exceed $3.049 \times 10^{-3}$, and as such error bars are not reported as they are too small. We observe error increasing as a function of manifold dimension, which is due to accumulation of errors. 

\subsection{Peaks Manifold} \label{sec:results:peaks}
The manifolds we have examined thus far have analytical distance functions; however, in many scenarios, we do not have access to the analytical distance functions. Therefore, we further test the metric-constrained Eikonal solver by considering a non-convex manifold with multiple local extrema (Figure~\ref{fig:manifolds}c), exhibiting non-negligible scalar curvature; namely the peaks function~\citep{mathworksPeaksFunctionMATLAB}
\begin{equation}
  \iota: [x; y] \mapsto [x; y; 3 (1 - x)^2 e^{-x^2 - (y + 1)^2} - 10 \left(\tfrac{x}{5} - x^3 - y^5 \right) e^{-x^2-y^2} - \tfrac{1}{3} e^{-(x + 1)^2 - y^2}].
\end{equation}
We first consider single-point solutions to the distance function, as described in §\ref{sec:methodology:sp_solutions}, obtaining a solution around the point $p = 0$. Next, we extend this to examine the global solution, as described in §\ref{sec:methodology:global_solutions}, comparing results with a cubic-spline-based length-minimising approach~\citep{stochman2021github}.

\subsubsection{Single-point solutions}
\begin{figure}[htb]
  \centering
  \begin{subfigure}[t]{0.30\textwidth}
    \centering
    \includegraphics[height=1\linewidth]{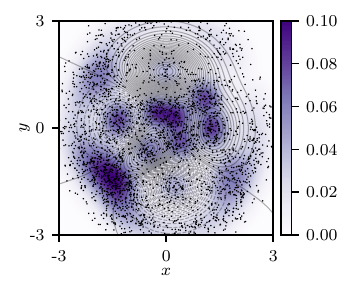}
    \caption{Curvature sampling}
    \label{fig:single_point:curvature_sampling}
  \end{subfigure} \hfill
  \begin{subfigure}[t]{0.30\textwidth}
    \centering
    \includegraphics[height=1\linewidth]{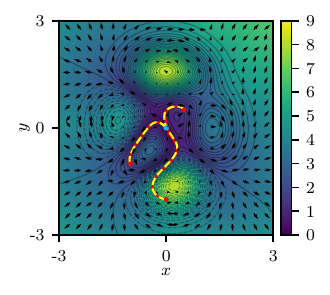}
    \caption{Distance field}
    \label{fig:single_point:geodesics}
  \end{subfigure} \hfill
  \begin{subfigure}[t]{0.30\textwidth}
    \centering
    \includegraphics[height=1\linewidth]{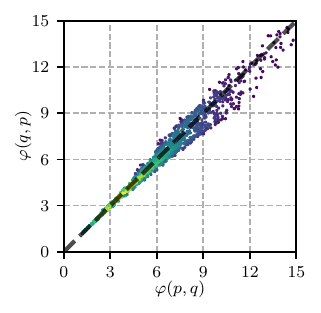}
    \caption{Symmetricity test}
    \label{fig:single_point:distance}
  \end{subfigure}
  \caption{Results for the single-point solution on the peaks manifold. Panel (\subref*{fig:single_point:curvature_sampling}) shows the probability density function generated from the curvature-based sampling (Algorithm~\ref{applications:algorithm:sampling}), with samples from the distribution overlaid. Panel (\subref*{fig:single_point:geodesics}) shows the distance field around a point in the centre of the domain with arrows depicting the geodesic flow overlaid. The yellow curves denote length-minimising geodesics obtained from the geodesic flow, while the red curves denotes the corresponding valid geodesics. Panel (\subref*{fig:single_point:distance}) shows results for the symmetricity test.}
\end{figure}

We first consider single-point solutions to the metric-constrained Eikonal equation. For optimisation, we use the \texttt{adam} optimiser with exponential weight decay; details of which can be found in the Appendix~\ref{app:training_parameters}. We train for a total of $2\times 10^5$ parameter updates, drawing $2 \times 10^{14}$ random samples from the domain for each parameter update. In this case, we employ the curvature-based sampling method described in §\ref{sec:methodology:network}, drawing random samples from the average of a uniform distribution and the distribution defined by the probability density function (PDF) of the Ricci scalar across the manifold. The curvature-based sampling is shown in Figure~\ref{fig:single_point:curvature_sampling}, with the colour map corresponding the the probability density function, and points representing random samples drawn from the distribution.
Results in Figure~\ref{fig:single_point:geodesics} show the predicted distance field from the centre of the domain $p$, with arrows representing the geodesic flow, $\grad \varphi_\theta$, overlaid. We observe that the geodesic distance adheres to the geometry. Using the predicted geodesic flow, we obtain length-minimising curves (yellow); these are compared with valid geodesics (red), generated through integration of the geodesic equation \eqref{eqn:geodesic_equation} with $\gamma(0) = q, \dot{\gamma}(0) = - \grad \varphi_\theta(q)$. The overlap of these two trajectories demonstrates the validity of the obtained geodesic flow.
Finally, to assess how the choice of $p \in M$ affects the performance, we investigate the symmetricity of the distance function across the domain. We consider a $7 \times 7$ grid of origin points on the manifold, distributed equally in the Euclidean sense, forming the set $\mathcal{X}$. For each $p, q \in \mathcal{X}$, we obtain a distance function and compute pair-wise distances, $\varphi_\theta(q; p), \varphi_\theta(p; q)$, as shown in Figure~\ref{fig:single_point:distance}. We observe that as the distance increases, the variability in predicted distance increases correspondingly as distance predictions accumulate error with increasing distance.

\subsubsection{Global Solutions} \label{sec:results:peaks:global}
Obtaining a solution constrained to a single point has limited uses. In general, we wish to obtain a distance function defined over the entire manifold. Analysis for the single-point case demonstrates that, although we obtain accurate results, we cannot guarantee symmetricity. Moving to a {\it global} solution, we can ensure that we satisfy the condition {\bf M3}.
In absence of an analytical solution for arbitrary manifolds, including the peaks manifold, we compare the metric-constrained Eikonal approach with the standard approach found in the literature (optimising parameterised curves~\citep{stochman2021github, chen2018MetricsDeepGenerative}. For the standard approach, we represent geodesics as cubic-splines. These cubic splines are defined in such a manner that the boundary conditions are enforced automatically. The curves are parameterised in such a manner that any parameterisation yields a valid curve. 
We optimise the parameters of the cubic splines through a gradient-descent based approach, seeking to minimise the energy of the resulting curve. Optimisation is  performed with a gradient-descent approach, using the \texttt{adam} optimiser with a learning rate of $10^{-3}$ for a total of $50\times10^3$ iterations. Convergence is achieved when the absolute magnitude of the gradient updates do not exceed $10^{-4}$.

\begin{figure}[htb]
  \centering
  \includegraphics[height=0.4\linewidth]{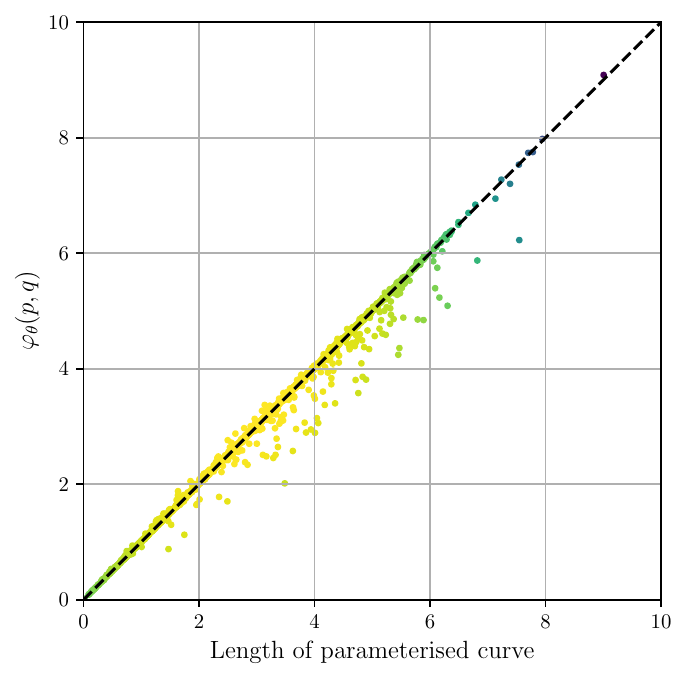}
  \caption{Comparison of results from standard parameterised curve approach against results from the metric-constrained Eikonal solver. Results are shown only in the cases where the parameterised curve approach converges. We observe good agreement between the proposed metric-constrained Eikonal approach, and the standard length-minimising curve approach.}
  \label{fig:global:distance_comparison}
\end{figure}

We employ the network $\varphi_\theta$ to represent the global distance function. This network is trained via the augmentation defined in Eq.\eqref{eqn:augmentation:global}, and results compared with those obtained from the parameterised curve approach. Results for this comparison are shown in Figure~\ref{fig:global:distance_comparison}. In many cases, the parameterised curve approach fails to converge, and so comparisons are shown only in cases where convergence was achieved. We observe that the Eikonal-based distances tend to be equal to, or less than, the lengths of the parameterised curves. This shows that the parameterised curves provide an upper-bound of the distance.

\begin{figure}[!h]
  \centering
  \includegraphics[width=\linewidth]{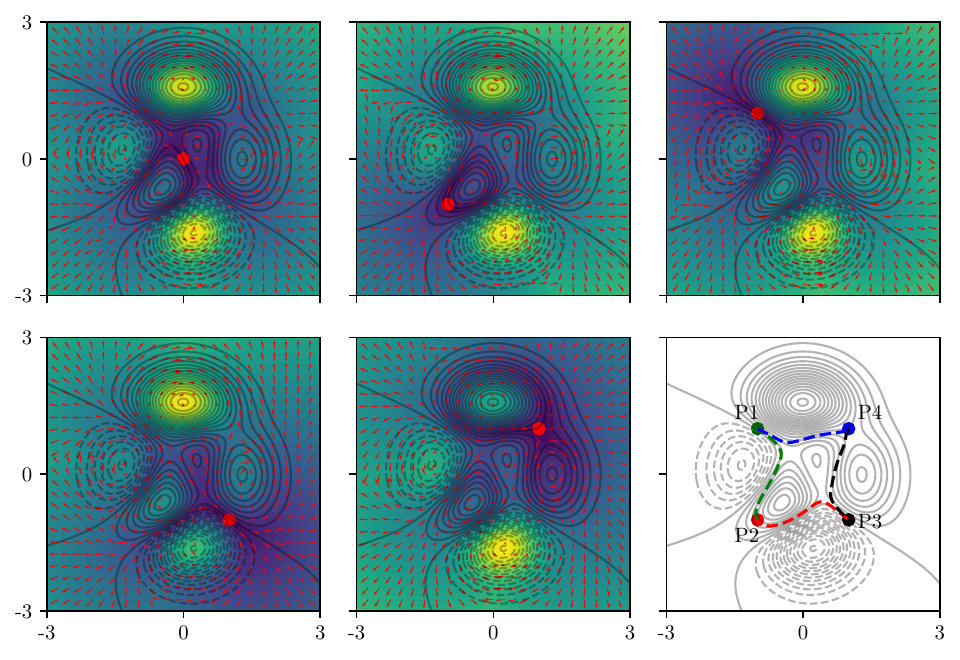}
  \caption{Distance fields and geodesic flows on the peaks manifold. The first five panels show a colour map of the geodesic distance with the vector field from the geodesic flow overlaid. The final panel in the lower right shows four geodesics generated from these vector fields, completing a cycle: P1 -- P4.}
  \label{fig:peaks:vector_fields}
\end{figure}

Given a point $p \in M$, we can visualise the resulting distance field around the point. Examples of this are provided in Figure~\ref{fig:peaks:vector_fields} where distance fields are shown for five distinct points. In addition to the distance fields, the geodesic flow is denoted by the red vector fields in the figure. The geodesic flow is defined across the entire domain, and can be employed to obtain globally length-minimising geodesics. In the bottom right panel of Figure~\ref{fig:peaks:vector_fields}, we take four of the points, labelled P1 -- P4, and obtain the geodesics connecting each point to its neighbours. We observe that we obtain a complete cycle with each geodesic connecting the points successfully.

\section{Applications} \label{sec:applications}
There exist a wide range of applications for which manifold-based distance computations are useful. Given a differentiable representation of the distance function on the manifold, we are able to perform optimisation with respect to the distance. To this end, we examine two applications of differentiable distance functions on manifolds: {\it (i)} computing the Fr\'echet mean~\citep{guigui2023IntroductionRiemannianGeometry}; and {\it (ii)} unsupervised clustering~\citep{jin2010KMeansClustering}. 

\subsection{Computing the Fr\'echet mean} \label{sec:frechet_mean}

The arithmetic mean is broadly incompatible with arbitrary manifolds, being suitable for operating in Euclidean spaces only~\citep{pennec2006IntrinsicStatisticsRiemannian, guigui2023IntroductionRiemannianGeometry}. To provide an illustrative example, consider two distinct points on the surface of $\mathbb{S}^2$ Taking the arithmetic mean of these points yields a mean interior to the sphere, rather than on the surface. The Fr\'echet mean accounts for this and is defined as the point on the manifold which minimises the squared distance to samples on the manifold, where the distance function is defined on the manifold~\citep{pennec2006IntrinsicStatisticsRiemannian}. The Fr\'echet mean is 
\begin{equation} \label{eqn:frechet_mean}
  \mu^\ast = \argmin_{\mu} \sum_{q \in \mathcal{X}} d_g (\mu, q)^2,
\end{equation}
where $\mathcal{X}$ denotes the set of points on the manifold for which we wish to compute the mean. Standard approaches for computing the Fr\'echet mean rely on computing the $\log$ map between pairs of points, ultimately requiring solving sub-optimisation problems, which becomes computationally expensive for large numbers of points. This standard approach is outlined in algorithm~\ref{applications:algorithm:frechet_mean}. Instead, we propose to first obtain a differentiable representation of the distance function using a metric-constrained Eikonal solver, outlined in algorithm~\ref{applications:algorithm:frechet_mean_new}. The obtained distance function provides fast inference, and allows us to optimise directly for the Fr\'echet mean. 
To provide a quantitative example, we must first define a manifold for which we can analytically compute the mean. For the purposes of this work, we employ a Gaussian mixture model.
\begin{algorithm}
\caption{Standard algorithm for computing Fr\'echet mean}\label{applications:algorithm:frechet_mean}
\begin{algorithmic}
\Require $q \in \mathcal{X} \subset M$
\State $\mu_0 \gets \tfrac{1}{\lvert \mathcal{X} \rvert} \sum_{p \in \mathcal{X}} p$
\For{$i \gets 1, \dots n$}
  \For{$p \in \mathcal{X}$}
    \State $\gamma_p \gets \verb|compute_geodesic|(\mu_{i-1}, p)$
    \State $\bm{L}_p \gets \tfrac{\dot{\gamma}_p (0)}{\lVert \dot{\gamma}_p (0) \rVert} L(\gamma_p)$
  \EndFor
  \State $\hat{\mu}_i \gets \tfrac{1}{\lvert \mathcal{X} \rvert} \sum_{x \in \bm{L}} x$
  \Comment{compute tangent space mean}
  \State $\mu_i \gets \exp_{\mu_{i-1}}(\hat{\mu}_i)$
  \Comment{map tangent space mean back to the feature space}
\EndFor
\end{algorithmic}
\end{algorithm}
~
\begin{algorithm}
\caption{Computing the Fr\'echet mean with the metric-constrained Eikonal solver}\label{applications:algorithm:frechet_mean_new}
\begin{algorithmic}
\Require $q \in \mathcal{X} \subset M$
\State $\varphi_\theta \gets \verb|solve_metric_constrained_eikonal|()$
\State $\mu_0 \gets \tfrac{1}{\lvert \mathcal{X} \rvert} \sum_{p \in \mathcal{X}} p$
\For{$i \gets 1, \dots n$}
  \State $\tau \gets \Sigma \varphi_\theta(\mu_{i - 1}, p)^2$
  \Comment{compute action to minimise}
  \State $\mu_i \gets \verb|gradient_step|(\mu_{i - 1}, \nicefrac{\partial \tau}{\partial \mu_{i - 1}})$
  \Comment{conduct gradient-based update}
\EndFor
\end{algorithmic}
\end{algorithm}

\subsubsection{Gaussian mixture models as a manifold}
The probability density function (PDF) of a Gaussian mixture model is a weighted sum of the PDFs of the component Gaussians
\begin{equation} \label{eqn:gmm_pdf}
  p(x; \mu, \Sigma, \alpha) = \Sigma_{i=1}^{N_G} \alpha_i f_{\mathcal{N}}(x; \mu_i, \Sigma_i) 
    \quad \text{s.t.} \quad \sum_{i=1}^{N_G} \alpha_i = 1
    \quad \text{with} \quad x, \mu_i \in \mathbb{R}^n, \Sigma_i \in \mathbb{R}^{n \times n}
\end{equation}
where for $N_G$ component Gaussians,  $\alpha_i$ is a weighting term, $\mu_i$ is the mean of the $i$th Gaussian, $\Sigma_i$ is the covariance matrix of the $i$th Gaussian, and $f_{\mathcal{N}}$ is the density function for a Gaussian distribution. Given the definition of a Gaussian mixture model, the mean is defined as the weighted sum of the constituent means, that is $\mu = \sum_{i=1}^{N_G} \alpha_i \mu_i$.
By using the Gaussian mixture model, we define a manifold as the surface of the resulting probability density function. We define an immersion $\iota: \mathbb{R}^{n} \rightarrow \mathbb{R}^{n + 1}$ as
\begin{equation}
  \iota: x \mapsto [x; p(x; \mu, \Sigma, \alpha)].
\end{equation}

For the applications explored in this paper, we consider a manifold of dimension $n=2$. In combination, the manifold and the corresponding, yet to be found, distance function define our metric space. The manifold employed for the Fr\'echet mean case can be seen in Figure~\ref{fig:manifolds:gmm}.

\subsubsection{Fr\'echet mean of the Gaussian Mixture Model}
\begin{figure}[htb]
  \centering
  \begin{subfigure}[t]{0.40\textwidth}
    \centering
    \includegraphics[height=1\linewidth]{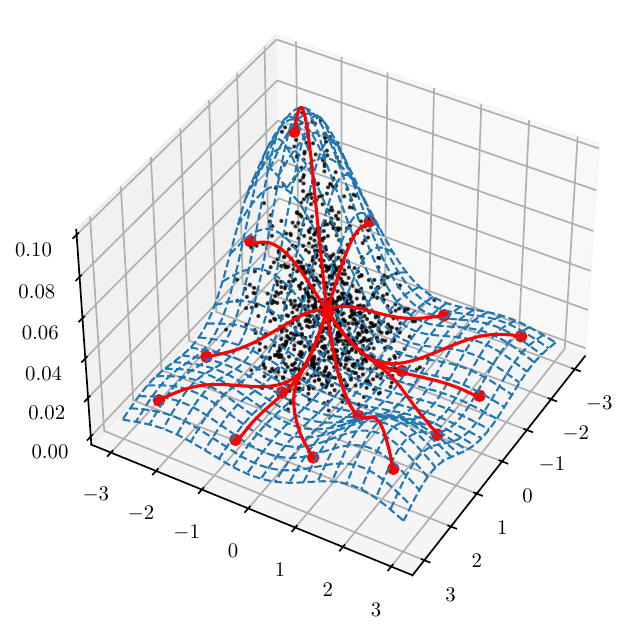}
    \caption{}
    \label{fig:frechet:extrinsic}
  \end{subfigure}
  \begin{subfigure}[t]{0.40\textwidth}
    \centering
    \includegraphics[height=1\linewidth]{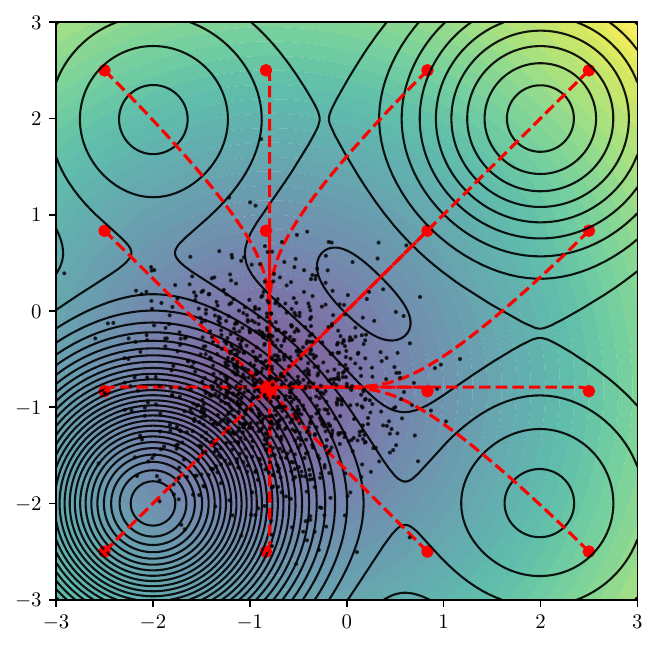}
    \caption{}
    \label{fig:frechet:intrinsic}
  \end{subfigure} \hfill
  \caption{Computing the Fr\'echet mean. Panel (\subref*{fig:frechet:extrinsic}) provides an extrinsic view of the manifold, the samples, and the optimisation trajectories -- each converging to the mean of the underlying distribution. Panel (\subref*{fig:frechet:intrinsic}) provides an intrinsic view of the same result. The arithmetic mean does not lie on the surface of the manifold, but it is interior to the surface.}
  \label{fig:frechet_mean}
\end{figure}

We first define the manifold as the surface of a Gaussian mixture model with four constituent Gaussians. The domain is then bounded such that $p \in [-3, 3] \subset \mathbb{R}^2$, which allows us to standardise inputs to the network. Parameters of the network are trained for a total of $5 \times 10^4$ parameter updates, using the \texttt{adam} optimiser to update the weights. The optimiser has an initial learning rate $\eta = 10^{-3}$, which decays exponentially. Further information on the network and optimiser can be found in Appendix~\ref{app:training_parameters}.
Using the same Gaussian mixture model as that used in training, we sample points on the surface of the manifold, discarding points that lie outside of the prescribed domain. The mean of these points is by definition the expected value of the given Gaussian mixture model. Given these points, and the distance function, we seek to recover the mean of the distribution by way of computing the Fr\'echet mean.
We compute the Fr\'echet mean through gradient-based optimisation, which is possible as a result of the differentiable nature of the distance function. We provide an initial guess $\mu_0$, and use the \texttt{adam} optimiser with learning rate $\eta = 10^{-2}$ to update the candidate mean. In practice, given the fast inference afforded by the distance function, we can run multiple optimisation chains in parallel. We provide sixteen initial guesses and optimise in parallel.
Figure \ref{fig:frechet_mean} shows the manifold, the samples, and the optimisation trajectories from each of the initial conditions. We observe that each of the optimisation trajectories terminates at the same point. For each these optimisation chains, we compute the relative $\ell^2$ error between the true mean, and the predicted mean. The mean of these errors is $\bar{\varepsilon} = 4.873 \times 10^{-3}$, showing we are able to accurately recover the Fr\'echet mean on the manifold defined by a Gaussian mixture model.

\subsection{Unsupervised clustering on the manifold with the metric-constrained Eikonal solver}
Clustering-based methods follow a similar principle to that of computing the Fr\'echet mean. Given a set of points $\mathcal{X}$, we seek to identify centroids and their associated clusters. For $k$-means clustering~\citep{lloyd1982least}, we define the objective
\begin{equation}
  \left(\mu^\ast, \mathcal{S}^\ast \right) = \argmin_{(\mu, \mathcal{S})} \sum_{i=1}^k \sum_{x \in S_i} d_g(x, \mu)^2 ,
\end{equation}
where for $k$ clusters, $\mu_i$ denotes the centroid of each cluster $\mathcal{S}_i$. To solve the optimisation problem, we make use of Lloyd's algorithm~\citep{lloyd1982least}, which alternates between two steps: {\it (i)} {assignment}, which assigns each observation to the cluster with the nearest centroid; and {\it (ii)} {update}, which recomputes the centroids of each cluster based on the observations assigned to the cluster. 

A large number of distance computations are required for convergence. In the assignment stage a total of $k\lvert \mathcal{X} \rvert$ distance computations are required, and the update stage relies on computing the Fr\'echet mean for each cluster as described in §\ref{sec:frechet_mean}. Existing methods are ill-equipped to solve such a problem: {\it (i)} the large number of distance computations required at each iteration is prohibitive; and {\it (ii)} the nested optimisations required for obtaining Fr\'echet means at each iteration is infeasible.



\begin{algorithm}
\caption{Unsupervised clustering on the manifold}\label{applications:algorithm:clustering}
\begin{algorithmic}
\Require $(\mu^{(0)}, S^{(0)})$
\Comment{assign initial guesses for clusters and their centroids}
\State $i \gets 0$
\State $\varphi_\theta \gets \verb|solve_metric_constrained_eikonal|()$
\While{$S^{(i)} \neq S^{(i - 1)}$}
  \State $d_{\mu \rightarrow p} \gets \varphi_\theta(\mu^{(i - 1)}, p)$
  \Comment{compute distance between each centroid and each data point}
  \State $S^{(i)} \gets \verb|assign_clusters|(d_{\mu \rightarrow p}) $
  \Comment{assign points to centroids that minimise the distance}
  \State $\mu^{(i)} \gets \verb|frechet_mean|(S^{(i)})$
  \Comment{recompute centroids of each cluster, as per algorithm~\ref{applications:algorithm:frechet_mean_new}}
  \State $i \gets i + 1$
\EndWhile
\end{algorithmic}
\end{algorithm}

%

\subsubsection{Clustering on the Peaks manifold}
\begin{figure}[htb]
  \centering
  \begin{subfigure}[t]{0.31\textwidth}
    \centering
    \includegraphics[height=1\linewidth]{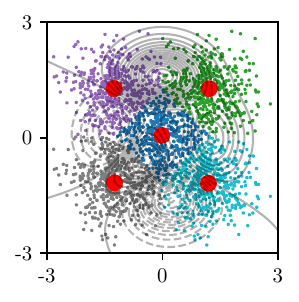}
    \caption{}
    \label{fig:clustering:peaks:euclidean}
  \end{subfigure}
  \begin{subfigure}[t]{0.31\textwidth}
    \centering
    \includegraphics[height=1\linewidth]{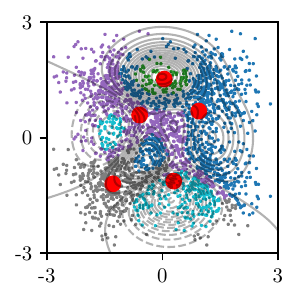}
    \caption{}
    \label{fig:clustering:peaks:euclidean_pullback}
  \end{subfigure}
  \begin{subfigure}[t]{0.31\textwidth}
    \centering
    \includegraphics[height=1\linewidth]{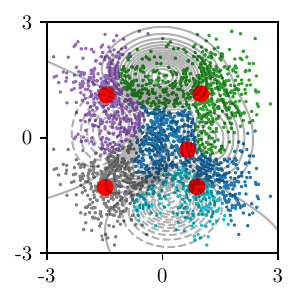}
    \caption{}
    \label{fig:clustering:peaks:ours}
  \end{subfigure} \hfill
  \caption{{Clustering on the peaks manifold. Panel (\subref*{fig:clustering:peaks:euclidean}) shows euclidean-based clustering of the samples directly on the manifold. Panel (\subref*{fig:clustering:peaks:euclidean_pullback}) shows the equivalent clustering, where distance is computed with the pullback of the Euclidean $\ell^2$ norm. Panel (\subref*{fig:clustering:peaks:ours}) shows the clustering results for the metric-constrained Eikonal approach. Red points, in each case, depict the centroids of the clusters.}}
  \label{fig:clusterering:peaks}
\end{figure}

We demonstrate the feasibility of this approach for points randomly sampled on the surface of the peaks manifold. First, samples $\mathcal{X}$ are generated on the surface of the manifold. We then use the previously obtained global distance function from §\ref{sec:results:peaks:global} in conjunction with Lloyd's algorithm to obtain a total of five distinct clusters. 
Results for this clustering are shown in Figure~\ref{fig:clusterering:peaks}. We first provide a baseline by conducting standard {\it k-}means clustering with a Euclidean distance function on the manifold, results for which are shown in Figure~\ref{fig:clustering:peaks:euclidean}. As a result of using the $\ell^2$ norm as the distance measure, the clusters form circles around each of the centroids. In Figure~\ref{fig:clustering:peaks:euclidean_pullback}, we extend the Euclidean-based clustering, using the pullback of the $\ell^2$ norm as the distance metric used for clustering, i.e, clustering in three-dimensional space. We observe that clusters are disjointed and overlapping; a characteristic that is unbecoming of clustering methods. Finally, the proposed clustering method is shown in Figure~\ref{fig:clustering:peaks:ours}. We  observe that points are clustered based on geodesic distance, conforming to the geometry of the manifold. This is particularly noticeable for the dark-blue cluster in the lower right which extends into the centre of the domain.
The deployment of the metric-augmented Eikonal solver in Figure~\ref{fig:clustering:peaks:ours} exemplifies the improvements afforded by the proposed methodology. Thanks to the differential nature of the obtained distance function, the Fr\'echet mean, as shown in §\ref{fig:frechet_mean}, can be obtained with ease and for relatively low computational expense. Notably, this example seeks to obtain five clusters over $2500$ samples. A total of $20$ iterations were required to reach convergence for the clustering, comprising $100$ evaluations of the Fr\'echet mean. This took just $113.084s$ of wall-clock time on a single NVIDIA GeForce RTX 4090.


\section{Conclusion} \label{sec:conclusion}

In this work, we address the problem of computing geodesic distances on manifolds. First, we introduce the metric-constrained Eikonal equation as a means for representing distance. Second, we propose a model-based parameterisation of the distance function providing a continuous, differentiable representation with theoretical guarantees. Obtaining the distance function in this manner is the main contribution of the manuscript; ultimately, alleviating issues encountered with all methods currently described in the literature. Third, we demonstrate the efficacy of our method for obtaining distance functions on a range of manifolds. A comparison is made to traditional methods in both cases, with our manifold-based Eikonal solver providing faster inference, and upholding theoretic guarantees in in all cases. Finally, we tackle two key applications, which are the computation of the Fr\'echet mean and $k$-means clustering on  manifolds. These applications have been computationally infeasible until now as a result of the comparatively expensive methods currently available in the literature. This work opens opportunities for reduced-order modelling, statistical modelling on manifolds, and distance-based computations in general.

\enlargethispage{20pt}

\subsubsection*{Acknowledgments}
D.K. is supported  by a PhD studentship sponsored by the Aeronautics department at Imperial College London. L.M. acknowledges support from the ERC Starting Grant No. PhyCo 949388 and EU-PNRR YoungResearcher TWIN ERC-PI 0000005. The authors thank Prof. Marika Taylor (Pro-Vice Chancellor and Head of College of Engineering and Physical Sciences at the University of Birmingham) for fruitful discussions on manifolds and geodesics.


\bibliography{references}

\begin{thebibliography}{37}
\providecommand{\natexlab}[1]{#1}
\providecommand{\url}[1]{\texttt{#1}}
\expandafter\ifx\csname urlstyle\endcsname\relax
  \providecommand{\doi}[1]{doi: #1}\else
  \providecommand{\doi}{doi: \begingroup \urlstyle{rm}\Url}\fi

\bibitem[Bradbury et~al.(2018)Bradbury, Frostig, Hawkins, Johnson, Leary, Maclaurin, Necula, Paszke, VanderPlas, {Wanderman-Milne}, and Zhang]{jax2018github}
James Bradbury, Roy Frostig, Peter Hawkins, Matthew~James Johnson, Chris Leary, Dougal Maclaurin, George Necula, Adam Paszke, Jake VanderPlas, Skye {Wanderman-Milne}, and Qiao Zhang.
\newblock {{JAX}}: Composable transformations of {{Python}}+{{NumPy}} programs, 2018.

\bibitem[Chen et~al.(2018)Chen, Klushyn, Kurle, Jiang, Bayer, and {van der Smagt}]{chen2018MetricsDeepGenerative}
Nutan Chen, Alexej Klushyn, Richard Kurle, Xueyan Jiang, Justin Bayer, and Patrick {van der Smagt}.
\newblock Metrics for {{Deep Generative Models}}, February 2018.

\bibitem[Crane et~al.(2013)Crane, Weischedel, and Wardetzky]{crane2013GeodesicsHeatNew}
Keenan Crane, Clarisse Weischedel, and Max Wardetzky.
\newblock Geodesics in heat: {{A}} new approach to computing distance based on heat flow.
\newblock \emph{ACM Transactions on Graphics}, 32\penalty0 (5):\penalty0 1--11, September 2013.
\newblock ISSN 0730-0301, 1557-7368.
\newblock \doi{10.1145/2516971.2516977}.

\bibitem[Detlefsen et~al.(2021)Detlefsen, Pouplin, Feldager, Geng, Kalatzis, Hauschultz, {Gonz{\'a}lez-Duque}, Warburg, Miani, and Hauberg]{stochman2021github}
Nicki~S. Detlefsen, Alison Pouplin, Cilie~W. Feldager, Cong Geng, Dimitris Kalatzis, Helene Hauschultz, Miguel {Gonz{\'a}lez-Duque}, Frederik Warburg, Marco Miani, and S{\o}ren Hauberg.
\newblock {{StochMan}}.
\newblock \emph{GitHub. Note: https://github.com/MachineLearningLifeScience/stochman/}, 2021.

\bibitem[Dijkstra(1959)]{dijkstra1959NoteTwoProblems}
E.~W. Dijkstra.
\newblock A note on two problems in connexion with graphs.
\newblock \emph{Numerische Mathematik}, 1\penalty0 (1):\penalty0 269--271, December 1959.
\newblock ISSN 0945-3245.
\newblock \doi{10.1007/BF01386390}.

\bibitem[do~Carmo(2013)]{carmo2013RiemannianGeometry}
Manfredo~Perdig{\~a}o do~Carmo.
\newblock \emph{Riemannian Geometry}.
\newblock Mathematics: Theory \& Applications. Birkh{\"a}user, Boston Basel Berlin, 14., corrected print edition, 2013.
\newblock ISBN 978-0-8176-3490-2 978-3-7643-3490-1.

\bibitem[Evans(2010)]{evans2010PartialDifferentialEquations}
Lawrence~C. Evans.
\newblock \emph{Partial {{Differential Equations}}}.
\newblock American Mathematical Soc., 2010.
\newblock ISBN 978-0-8218-4974-3.

\bibitem[Fefferman et~al.(2016)Fefferman, Mitter, and Narayanan]{fefferman2016TestingManifoldHypothesisa}
Charles Fefferman, Sanjoy Mitter, and Hariharan Narayanan.
\newblock Testing the manifold hypothesis.
\newblock \emph{Journal of the American Mathematical Society}, 29\penalty0 (4):\penalty0 983--1049, October 2016.
\newblock ISSN 0894-0347, 1088-6834.
\newblock \doi{10.1090/jams/852}.

\bibitem[Gao et~al.(2023)Gao, Xiong, Gao, Jia, Pan, Bi, Dai, Sun, and Wang]{gao2023retrieval}
Yunfan Gao, Yun Xiong, Xinyu Gao, Kangxiang Jia, Jinliu Pan, Yuxi Bi, Yi~Dai, Jiawei Sun, and Haofen Wang.
\newblock Retrieval-augmented generation for large language models: {{A}} survey.
\newblock \emph{arXiv preprint arXiv:2312.10997}, 2023.

\bibitem[Gropp et~al.(2020)Gropp, Yariv, Haim, Atzmon, and Lipman]{gropp2020ImplicitGeometricRegularization}
Amos Gropp, Lior Yariv, Niv Haim, Matan Atzmon, and Yaron Lipman.
\newblock Implicit {{Geometric Regularization}} for {{Learning Shapes}}, July 2020.

\bibitem[Grubas et~al.(2023)Grubas, Duchkov, and Loginov]{grubas2023NeuralEikonalSolver}
Serafim Grubas, Anton Duchkov, and Georgy Loginov.
\newblock Neural {{Eikonal}} solver: {{Improving}} accuracy of physics-informed neural networks for solving eikonal equation in case of caustics.
\newblock \emph{Journal of Computational Physics}, 474:\penalty0 111789, February 2023.
\newblock ISSN 0021-9991.
\newblock \doi{10.1016/j.jcp.2022.111789}.

\bibitem[Guigui et~al.(2023)Guigui, Miolane, and Pennec]{guigui2023IntroductionRiemannianGeometry}
Nicolas Guigui, Nina Miolane, and Xavier Pennec.
\newblock Introduction to {{Riemannian Geometry}} and {{Geometric Statistics}}: From basic theory to implementation with {{Geomstats}}.
\newblock \emph{Foundations and Trends{\textregistered} in Machine Learning}, 16\penalty0 (3):\penalty0 329--493, 2023.

\bibitem[Hastings(1970)]{hastings1970MonteCarloSampling}
W.~K. Hastings.
\newblock Monte {{Carlo Sampling Methods Using Markov Chains}} and {{Their Applications}}.
\newblock \emph{Biometrika}, 57\penalty0 (1):\penalty0 97--109, 1970.
\newblock ISSN 0006-3444.
\newblock \doi{10.2307/2334940}.

\bibitem[Hornik et~al.(1989)Hornik, Stinchcombe, and White]{hornik1989MultilayerFeedforwardNetworks}
Kurt Hornik, Maxwell Stinchcombe, and Halbert White.
\newblock Multilayer feedforward networks are universal approximators.
\newblock \emph{Neural Networks}, 2\penalty0 (5):\penalty0 359--366, January 1989.
\newblock ISSN 0893-6080.
\newblock \doi{10.1016/0893-6080(89)90020-8}.

\bibitem[Huckemann \& Ziezold(2006)Huckemann and Ziezold]{huckemann2006PrincipalComponentAnalysis}
Stephan Huckemann and Herbert Ziezold.
\newblock Principal component analysis for {{Riemannian}} manifolds, with an application to triangular shape spaces.
\newblock \emph{Advances in Applied Probability}, 38\penalty0 (2):\penalty0 299--319, June 2006.
\newblock ISSN 0001-8678, 1475-6064.
\newblock \doi{10.1239/aap/1151337073}.

\bibitem[Jin \& Han(2010)Jin and Han]{jin2010KMeansClustering}
Xin Jin and Jiawei Han.
\newblock K-{{Means Clustering}}.
\newblock In Claude Sammut and Geoffrey~I. Webb (eds.), \emph{Encyclopedia of {{Machine Learning}}}, pp.\  563--564. Springer US, Boston, MA, 2010.
\newblock ISBN 978-0-387-30164-8.
\newblock \doi{10.1007/978-0-387-30164-8\_425}.

\bibitem[Kimmel \& Sethian(1998)Kimmel and Sethian]{kimmel1998ComputingGeodesicPaths}
R.~Kimmel and J.~A. Sethian.
\newblock Computing geodesic paths on manifolds.
\newblock \emph{Proceedings of the National Academy of Sciences}, 95\penalty0 (15):\penalty0 8431--8435, July 1998.
\newblock ISSN 0027-8424, 1091-6490.
\newblock \doi{10.1073/pnas.95.15.8431}.

\bibitem[Kingma \& Ba(2017)Kingma and Ba]{kingma2017AdamMethodStochastic}
Diederik~P. Kingma and Jimmy Ba.
\newblock Adam: {{A Method}} for {{Stochastic Optimization}}, January 2017.

\bibitem[Lee(2012)]{lee2012IntroductionSmoothManifolds}
John~M. Lee.
\newblock \emph{Introduction to {{Smooth Manifolds}}}, volume 218 of \emph{Graduate {{Texts}} in {{Mathematics}}}.
\newblock Springer New York, New York, NY, 2012.
\newblock ISBN 978-1-4419-9981-8 978-1-4419-9982-5.
\newblock \doi{10.1007/978-1-4419-9982-5}.

\bibitem[Lee(2018)]{lee2018IntroductionRiemannianManifolds}
John~M. Lee.
\newblock \emph{Introduction to {{Riemannian Manifolds}}}, volume 176 of \emph{Graduate {{Texts}} in {{Mathematics}}}.
\newblock Springer International Publishing, Cham, 2018.
\newblock ISBN 978-3-319-91754-2 978-3-319-91755-9.
\newblock \doi{10.1007/978-3-319-91755-9}.

\bibitem[Lewis et~al.(2021)Lewis, Perez, Piktus, Petroni, Karpukhin, Goyal, K{\"u}ttler, Lewis, Yih, Rockt{\"a}schel, Riedel, and Kiela]{lewis2021RetrievalAugmentedGenerationKnowledgeIntensive}
Patrick Lewis, Ethan Perez, Aleksandra Piktus, Fabio Petroni, Vladimir Karpukhin, Naman Goyal, Heinrich K{\"u}ttler, Mike Lewis, Wen-tau Yih, Tim Rockt{\"a}schel, Sebastian Riedel, and Douwe Kiela.
\newblock Retrieval-{{Augmented Generation}} for {{Knowledge-Intensive NLP Tasks}}, April 2021.

\bibitem[Lin et~al.(2014)Lin, Yang, He, and Ye]{lin2014GeodesicDistanceFunction}
Binbin Lin, Ji~Yang, Xiaofei He, and Jieping Ye.
\newblock Geodesic {{Distance Function Learning}} via {{Heat Flow}} on {{Vector Fields}}, May 2014.

\bibitem[Lloyd(1982)]{lloyd1982least}
Stuart Lloyd.
\newblock Least squares quantization in {{PCM}}.
\newblock \emph{IEEE transactions on information theory}, 28\penalty0 (2):\penalty0 129--137, 1982.

\bibitem[Metropolis et~al.(1953)Metropolis, Rosenbluth, Rosenbluth, Teller, and Teller]{metropolis1953equation}
Nicholas Metropolis, Arianna~W Rosenbluth, Marshall~N Rosenbluth, Augusta~H Teller, and Edward Teller.
\newblock Equation of state calculations by fast computing machines.
\newblock \emph{The journal of chemical physics}, 21\penalty0 (6):\penalty0 1087--1092, 1953.

\bibitem[Ni \& Wang(2023)Ni and Wang]{ni2023ShapeAnalysisComputing}
Qian Ni and Xuhui Wang.
\newblock Shape {{Analysis}} by {{Computing Geodesics}} on a {{Manifold}} via {{Cubic B-splines}}.
\newblock \emph{Communications in Mathematics and Statistics}, October 2023.
\newblock ISSN 2194-671X.
\newblock \doi{10.1007/s40304-023-00373-3}.

\bibitem[Pennec(2006)]{pennec2006IntrinsicStatisticsRiemannian}
Xavier Pennec.
\newblock Intrinsic {{Statistics}} on {{Riemannian Manifolds}}: {{Basic Tools}} for {{Geometric Measurements}}.
\newblock \emph{Journal of Mathematical Imaging and Vision}, 25\penalty0 (1):\penalty0 127--154, July 2006.
\newblock ISSN 0924-9907, 1573-7683.
\newblock \doi{10.1007/s10851-006-6228-4}.

\bibitem[Radford et~al.(2021)Radford, Kim, Hallacy, Ramesh, Goh, Agarwal, Sastry, Askell, Mishkin, Clark, Krueger, and Sutskever]{radford2021LearningTransferableVisual}
Alec Radford, Jong~Wook Kim, Chris Hallacy, Aditya Ramesh, Gabriel Goh, Sandhini Agarwal, Girish Sastry, Amanda Askell, Pamela Mishkin, Jack Clark, Gretchen Krueger, and Ilya Sutskever.
\newblock Learning {{Transferable Visual Models From Natural Language Supervision}}.
\newblock In \emph{Proceedings of the 38th {{International Conference}} on {{Machine Learning}}}, pp.\  8748--8763. PMLR, July 2021.

\bibitem[Rajkovi{\'c}(2023)]{rajkovic2023GeodesicsSplinesEmbeddings}
Marko Rajkovi{\'c}.
\newblock \emph{Geodesics, {{Splines}}, and {{Embeddings}} in {{Spaces}} of {{Images}}}.
\newblock Thesis, Universit{\"a}ts- und Landesbibliothek Bonn, August 2023.

\bibitem[Sethian(1999)]{sethian1999FastMarchingMethods}
J.~A. Sethian.
\newblock Fast {{Marching Methods}}.
\newblock \emph{SIAM Review}, 41\penalty0 (2):\penalty0 199--235, 1999.
\newblock ISSN 0036-1445.

\bibitem[Varadhan(1967)]{varadhan1967BehaviorFundamentalSolution}
S.~R.~S. Varadhan.
\newblock On the behavior of the fundamental solution of the heat equation with variable coefficients.
\newblock \emph{Communications on Pure and Applied Mathematics}, 20\penalty0 (2):\penalty0 431--455, 1967.
\newblock ISSN 1097-0312.
\newblock \doi{10.1002/cpa.3160200210}.

\bibitem[Wald(2010)]{wald2010general}
Robert~M Wald.
\newblock \emph{General Relativity}.
\newblock University of Chicago press, 2010.

\bibitem[Wang et~al.(2021)Wang, Teng, and Perdikaris]{wang2021UnderstandingMitigatingGradient}
Sifan Wang, Yujun Teng, and Paris Perdikaris.
\newblock Understanding and {{Mitigating Gradient Flow Pathologies}} in {{Physics-Informed Neural Networks}}.
\newblock \emph{SIAM Journal on Scientific Computing}, 43\penalty0 (5):\penalty0 A3055--A3081, January 2021.
\newblock ISSN 1064-8275.
\newblock \doi{10.1137/20M1318043}.

\bibitem[Wang et~al.(2023)Wang, Sankaran, Wang, and Perdikaris]{wang2023ExpertGuideTraining}
Sifan Wang, Shyam Sankaran, Hanwen Wang, and Paris Perdikaris.
\newblock An {{Expert}}'s {{Guide}} to {{Training Physics-informed Neural Networks}}, August 2023.

\bibitem[Wang et~al.(2014)Wang, Huang, Wang, and Wang]{wang2014generalized}
Wei Wang, Yan Huang, Yizhou Wang, and Liang Wang.
\newblock Generalized autoencoder: {{A}} neural network framework for dimensionality reduction.
\newblock In \emph{Proceedings of the {{IEEE}} Conference on Computer Vision and Pattern Recognition Workshops}, pp.\  490--497, 2014.

\bibitem[Zhang \& Zha(2004)Zhang and Zha]{zhang2004principal}
Zhenyue Zhang and Hongyuan Zha.
\newblock Principal manifolds and nonlinear dimensionality reduction via tangent space alignment.
\newblock \emph{SIAM journal on scientific computing}, 26\penalty0 (1):\penalty0 313--338, 2004.

\bibitem[Zhao(2005)]{zhao2005fast}
Hongkai Zhao.
\newblock A fast sweeping method for eikonal equations.
\newblock \emph{Mathematics of computation}, 74\penalty0 (250):\penalty0 603--627, 2005.

\bibitem[Zhao(2007)]{zhao2007parallel}
Hongkai Zhao.
\newblock Parallel implementations of the fast sweeping method.
\newblock \emph{Journal of Computational Mathematics}, pp.\  421--429, 2007.

\end{thebibliography}
\bibliographystyle{tmlr}

\appendix

\section{Imposing upper bounds in network augmentation} \label{app:upper_bounds}

Considering points $p, q \in M$, there must exist no shorter curve connecting the points than the length-minimising geodesic, as described in Eq.~\eqref{eqn:review:distance}. Under this assumption, the length of an arbitrary curve $\gamma_L$ that connects the points must be greater than, or equal to, the length of the minimising geodesic.  We can prescribe the bounds
\begin{equation}
  (\iota^\ast d_E)(p, q) 
  \leq
    d(p, q)
  \leq
    \int_0^1 \langle {\dot \gamma_L}(\lambda), {\dot \gamma_L}(\lambda) \rangle_{\gamma_L (\lambda)} d\lambda.
\end{equation}
We can use these bounds to alter the way in which we augment the network outputs. Empirically, prescribing an upper-bound incurs additional computational cost with little improvement in performance. For this reason, we do not show results for upper-bounded augmentations in the paper. 
Choosing an arbitrary curve between two points prescribes this upper bound. In order to reliably obtain curves, we can linearly interpolate between points in the intrinsic coordinates, that is:
\begin{equation} 
  \gamma_L(t) = q + (p - q) t \qquad \text{where, } \quad t \in [0, 1],
\end{equation}
allowing us to compute the upper-bounded length as
\begin{equation} 
  L^{+}_{\gamma} = \int_0^1 \langle \dot{\gamma}_{L}(t), \dot{\gamma}_{L}(t) \rangle_{\gamma(t)}^{\nicefrac{1}{2}} dt
\end{equation}
or,
\begin{equation} 
  L^{+}_{\gamma} = \int_0^1 \langle p - q, p - q \rangle_{q + (p - q)t}^{\nicefrac{1}{2}} dt
\end{equation}
Using this definition, we can alter how we augment the network output to provide a bounded solution. For the single-point solution, we would write
\begin{equation}
  \varphi_\theta(q; p)
  = \left( \iota^\ast d_E \right)(p, q) \left[
    1 + \left( \frac{L^{+}_\gamma}{(\iota^\ast d_E) (p, q)} - 1 \right) (\sigma \circ \tilde{\varphi}) (q; p)
  \right],
\end{equation}
where $\sigma$ now represents the \texttt{sigmoid} function, and $L^{+}_\gamma$ is the length of the linearly interpolated curve in intrinsic coordinates.
We can extend this approach to the two-point solution 
\begin{equation}
  \varphi_\theta(p, q)
  = \left( \iota^\ast d_E \right)(p, q) \left[
    1 + \left( \frac{L^{+}_\gamma}{(\iota^\ast d_E) (p, q)} - 1 \right) \sigma \left( \tfrac{1}{2} (\tilde{\varphi}_\theta (p, q) + \tilde{\varphi}_\theta (q, p) ) \right)
  \right],
\end{equation}
With these approaches, we have retained the same guarantees as the network augmentations described in the main text, as well as provided an upper-bound for the distance.

\section{Training parameters} \label{app:training_parameters}

In each of the cases, we employ a modified MLP architecture. The architecture has a depth of $8$ layers, each with a width of $256$ neurons. Optimisation was conducted with the \texttt{adam} optimiser. The optimiser uses an exponentially decaying learning rate schedule, reducing by a factor $0.9$ every $2000$ parameter updates.

\section{Hypersphere} \label{app:hypersphere}

For the hypersphere, we define a coordinate system with $n - 1$ angular coordinates $x_1, ..., x_{n - 1}$, with angles $x_1, ..., x_{n - 2} \in [0, \pi]$, and $x_{n - 1} \in [0, 2\pi)$ radians. The immersion is the mapping $\iota : x \hookrightarrow y$, where
\begin{align} \label{eqn:hypersphere:mapping}
  y_1 =& \cos(x_1) \\
  y_2 =& \sin(x_1) \cos(x_2) \\
  y_3 =& \sin(x_1) \sin(x_2) \cos(x_3) \\
  & \vdots \\
  y_{n - 1} =& \sin(x_1) \dots \sin(x_{n - 2}) \cos(x_{n - 1}) \\
  y_n =& \sin(x_1) \dots \sin(x_{n - 2}) \sin(x_{n - 1}).
\end{align}

\end{document}